\documentclass[graybox]{svmult}
\usepackage[utf8]{inputenc}
\usepackage{booktabs}
\usepackage{hyperref}
\usepackage{mathptmx}
\usepackage{amsmath}
\usepackage{amsfonts}
\usepackage{helvet}
\usepackage{courier}
\usepackage{type1cm}
\usepackage{makeidx}
\usepackage{graphicx}
\usepackage{multicol}
\usepackage[bottom]{footmisc}
\usepackage{wrapfig}
\usepackage{textcomp}
\usepackage{tikz}

\usetikzlibrary{shapes,arrows,fit,decorations.pathreplacing,calligraphy}

\newcommand{\argmax}[1]{\underset{#1}{\operatorname{argmax}}\;}
\DeclareMathOperator{\sign}{sgn}
\makeindex

\begin{document}

\title*{Question-Answer Selection in User to User Marketplace Conversations}
\author{Girish Kumar, Matthew Henderson, Shannon Chan, Hoang Nguyen, Lucas Ngoo}
\institute{Carousell Pte. Ltd., 240 Tanjong Pagar Rd, Singapore}

\maketitle

\abstract{
Sellers in user to user marketplaces can be inundated with questions from potential buyers. Answers are often already available in the product description. We collected a dataset of around 590K such questions and answers from conversations in an online marketplace. We propose a question answering system that selects a sentence from the product description using a neural-network ranking model. We explore multiple encoding strategies, with recurrent neural networks and feed-forward attention layers yielding good results. This paper presents a demo to interactively pose buyer questions and visualize the ranking scores of product description sentences from live online listings.}

% lit review- (QA) current state of QA, squad, extractive vs generative, phrase vs sentence. (conversation) generative neural conversation models, vs ranking: smart reply
\section{Introduction}
In online marketplaces, buyers primarily deal with sellers through chats -- to ask questions, negotiate, and arrange logistics. Any improvement to these interactions can greatly increase user satisfaction and the effectiveness of the marketplace. We present a system that can automatically answer buyer questions by reading the product description, freeing the sellers' time and giving buyers quick answers.

Current question answering methods consist of generative and extractive techniques. For example, Yin et al. present a generative approach that conditions a word-level recurrent neural network (RNN) on facts retrieved from a knowledge base \cite{yin2015neural}. Others model facts using memory modules, which RNNs attend over \cite{sukhbaatar2015end, weston2014memory}.

Extractive techniques extract an answer sentence or phrase from a given context.
Many recent phrase-level extractive models are built and evaluated on the Stanford Question Answering Dataset (SQuAD) \cite{rajpurkar2016squad}. SQuAD provides human-generated questions along with the context and human-extracted answer spans. Much recent work has used neural networks that first encode the question and the corresponding context with word-level RNNs, followed by attention mechanisms over the context. Pointer networks, conditioned on the aligned representations, then point to the start and end indices of the predicted answer span, with respect to the context \cite{wang2016machine, seo2016bidirectional}.

Sentence-level extractive methods start by encoding questions and each candidate sentence in the context. Binary classifiers can be trained on the encodings to predict if a sentence answers the question, choosing the candidate sentence with the highest prediction score \cite{guo2017enhanced}. Masala et al. use a hinge loss to maximize the similarity between question-answer pairs and minimize otherwise \cite{masala}. 
%Wang et al. introduce attention mechanisms to the encoding process \cite{wang2016inner}.
We focus on sentence-level extractive techniques for their speed, to not greatly degrade chat latency.

Conversational reply methods are of interest since we deal with chats. Approaches include generative \emph{seq2seq} models- using a RNN to encode the previous message and another RNN to generate a response word-by-word \cite{shang2015neural, vinyals2015neural}. Henderson et al. propose a model to rank the dot product of the previous message vector and the actual response vector favourably against other response vectors in a training batch \cite{henderson2017efficient}. For efficiency reasons, the ranking approach greatly appeals to us.

In this paper, we present an extension of Henderson et al.'s dot product model for answer-sentence selection. We explore adding Long Short-Term Memory (LSTM) networks and attention mechanisms to the base model. Finally, we present a demo where a user can ask questions about a product listing and view model predictions.
\vspace{-0.2cm}
\section{Data}
\subsection{Data Collection}
\begin{wraptable}{l}{7.2cm}
\vspace{-0.7cm}
{\footnotesize
\begin{tabular}{p{3.2cm}@{\hskip 2.5mm} p{3.8cm}}
  \toprule
  \textbf{Chat} & \textbf{Description} \\ 
  \midrule
    \textbf{B}: Can you do delivery? \newline
    \textbf{S}: Yes, delivery is \$15. \newline
    \textbf{B}: Great. Is it sturdy? \newline
    \textbf{S}: Yes! It's well built. \newline
    \textbf{B}: What colours are there?
    &
    {\scriptsize This is one of the best cat towers we offer and your cats will love it. \newline
    At 185cm tall, it's a great vertical gym. \newline
    8 scratch posts ensure healthy nails. \newline
    You've a choice of two colours. \newline
    We sell it in cream-white or black. %\newline
    %Dimensions: 50 x 30 x 185 cm.
    }

    \\
  \textbf{Next message} & \textbf{Answer Sentence}  \\ \midrule
  We have cream-white or black. &
  We  sell  it  in  cream-white or black. \\
  \bottomrule
\end{tabular}
}
\caption{A dataset sample, giving a conversational context, a description whose sentences are treated as candidate answers, and the correct answer. The true next message is used to identify the correct answer. \textbf{B} \& \textbf{S} denote buyer \& seller.}
\vspace{-0.87cm}
\end{wraptable}
Carousell is an online used goods marketplace. Potential buyers deal with sellers through chats. A corpus of 36M of such chats, containing 400M messages, was available. This corpus is encrypted and engineers can only inspect aggregated statistics across many users. We specifically collect buyer questions with seller replies repeating phrases in the product description. The description sentence with the repeated phrase is taken as the answer sentence. We also store the conversation history, i.e. buyer \& seller messages till the seller's answer. 590K examples are collected.

The model also needs to identify whether or not the buyer's message has an answer in the product description. We collected 345K buyer messages without an answer in the description, with the dataset containing 935K examples in total.

% MH: mention train/test split somewhere
\vspace{-0.65cm}
\subsection{Evaluation Metrics}
\label{dataeval}

Consider a question, $q$, and $N$ candidate answer sentences, $\left(a_1,\:a_2,\:\ldots,\:a_N\right)$. We add a special token, $a_0$, to be predicted when there is no suitable answer, and write $\mathbf{A}=\left(a_0,\:a_1,\:\ldots,\:a_n \right)$. A question answering model must produce a distribution, $P_i$, over the $a_i$, where $P_i = P(a_i \ |\ \mathbf{A},\:q)$ and $\sum_{i=0}^{n} P_i = 1$.
Given $P_i$, and the labelled index of the correct answer sentence, $k$, we compute accuracy as follows. Note that $k=0$ when there is no suitable answer.
\vspace{-0.3cm}
\begin{equation}
  \text{Accuracy} =
  \begin{cases}
                                   1 & \text{if} \ \argmax{i} P_i = k \\
                                   0 & \text{otherwise}
  \end{cases}
 \vspace{-0.25cm}
\end{equation}
We compute the \emph{Overall Accuracy} averaged  over all the test samples. While our formulation requires the model to decide if $q$ has an answer and predict an answer sentence jointly, it is possible to train separate models for each task. Therefore, we calculate, separately, the \emph{Positive Accuracy} over test samples with $k > 0$. To evaluate how well the model predicts if there is an answer, we present a triggering accuracy:
\vspace{-0.2cm}
\begin{equation}
    \text{Trigger Accuracy} =   \begin{cases}
                                   1 & \text{if}\  \sign(\argmax{i} P_i) = \sign(k) \\
                                   0 & \text{otherwise}
  \end{cases}
\end{equation}
\vspace{-0.35cm}
\section{Model Architecture}
\label{modelsec}
% MH: cite the figure somewhere x

In this section, we describe an approach to obtain $P_i = P(a_i \ |\ \mathbf{A},\:q)$. As in \cite{henderson2017efficient}, we estimate $P_i$ using the dot product of two neural network functions, $\mathbf{h}(\cdot)$, $\mathbf{g}(\cdot)$:
\vspace{-0.2cm}
\begin{equation}
   P_i = P(a_i \ |\ \mathbf{A},\:q) \approx \frac{e^{\mathbf{h}(q)^{\text{T}}\mathbf{g}(a_i,\ \mathbf{A},\ q)}}{\sum_{j=0}^{N}{e^{\mathbf{h}(q)^{\text{T}}\mathbf{g}(a_j,\ \mathbf{A}, \ q)}}}
\end{equation}
\vspace{-0.2cm}

The softmax function ensures $\sum_{i=0}^{N} P_i = 1$. Separating the model into two networks allows the network to run efficiently on varying sizes of $\mathbf{A}$.
%The following subsections elaborate how $\mathbf{h}(\cdot)$ and  $\mathbf{g}(\cdot)$ encode $q$ and $\mathbf{A_+}$ respectively.
\subsection{N-gram Representation}

% MH: add details of the vocabulary, how many unigrams, how many bigrams? x

The $\mathbf{h}(\cdot)$ and  $\mathbf{g}(\cdot)$ sub-networks start by extracting n-gram features from $q$ and $\mathbf{A}$. Embeddings are learnt for each n-gram during training. For each question and answer sentence, the embeddings of their n-grams are summed. We denote this representation as $\psi(\cdot) \in \mathbb{R}^d$. A total of 100K unigrams and 200K bigrams were extracted from the full conversation corpus.
\vspace{-0.5cm}
\subsection{Encoding Techniques}
\begin{figure}
    \newcommand{\vertlstm}{\parbox{2mm}{\centering\tiny L\\S\\T\\M}}
    \centering
    \tikzset{%
      block/.style = {draw, thick, rectangle},
      vector/.style = {draw, thick, rectangle, rounded corners=3},
      input/.style = {},
      lstm/.style = {}
    }
    \begin{tikzpicture}[auto, node distance=10mm, >=stealth', scale=0.8, every node/.style={scale=0.8}]

        \node[input] (question) {$\psi(q)$};
        
        \node [input, right of=question, node distance=2cm] (a1) {$\psi(a_1)$};
        \node [input, right of=a1, xshift=3mm] (a2) {$\psi(a_2)$};
        \node [input, right of=a2] (ellipses) {$\ldots$};
        \node [input, right of=ellipses] (an) {$\psi(a_n)$};
        
        \node [input, left of=question, xshift=-4mm] (mh) {$\psi(m_H)$};
        \node [input, left of=mh] (ellipsesm) {$\ldots$};
        \node [input, left of=ellipsesm] (m2) {$\psi(m_2)$};
        \node [input, left of=m2] (m1) {$\psi(m_1)$};
        
        \node [block, lstm, above of=m1] (mlstm1) {\vertlstm};
        \node [block, lstm, above of=m2] (mlstm2) {\vertlstm};
        \node [right of=mlstm2] (ellipsesm2) {$\ldots$};
        \node [block, lstm, above of=mh] (mlstmh) {\vertlstm};

        \node [block, lstm, above of=question] (mlstmq) {\vertlstm};
        \node [vector, above of=mlstmq] (hq) {$\mathbf{h}$};
        
        \node [block, lstm, above of=a1] (lstm1) {\vertlstm};
        \node [block, lstm, above of=a2] (lstm2) {\vertlstm};
        \node [right of=lstm2] (ellipses2) {$\ldots$};
        \node [block, lstm, above of=an] (lstmn) {\vertlstm};

        \node [above of=lstm1] (attn1) {};
        \node [left of=attn1, xshift=2mm] (anchor1) {};
        \node [above of=lstm2] (attn2) {};
        \node [above of=ellipses2] (attnell) {};
        \node [above of=lstmn] (attnn) {};
        \node [right of=attnn, xshift=-2mm] (anchor2) {};
        \node [block, fit=(anchor1) (anchor2), inner sep=2mm] (attn) {self-attention};
        
        \node [vector, above of=attn1] (g1) {$\mathbf{g}_1$};
        \node [vector, above of=attn2] (g2) {$\mathbf{g}_2$};
        \node [right of=g2] (ellipses3) {$\ldots$};
        \node [vector, above of=attnn] (gn) {$\mathbf{g}_n$};
        \node [vector, left of=g1, xshift=-3mm] (g0) {$\mathbf{g}_0$};
        
        \node [vector, above of=g0] (hg0) {$\mathbf{h}^T\mathbf{g}_0$};
        \node [vector, above of=g1] (hg1) {$\mathbf{h}^T\mathbf{g}_1$};
        \node [vector, above of=g2] (hg2) {$\mathbf{h}^T\mathbf{g}_2$};
        \node [right of=hg2] (ellipses4) {$\ldots$};
        \node [vector, above of=gn] (hgn) {$\mathbf{h}^T\mathbf{g}_n$};
        
        \node [block, above of=hg2] (softmax) {softmax};
        \node [vector, right of=softmax, xshift=4mm, yshift=4mm] (p) {$\mathbf{P}$};
        
        \node [below of=m2, xshift=4mm, node distance=5mm] (messagesnote) {\small conversational context};
        \node [below of=question, node distance=5mm] (questionnote) {\small question};
        \node [below of=a2, xshift=4mm, node distance=5mm] (candidatenote) {\small candidate answers};
        
    \draw[decoration={brace, mirror}, decorate] (m1.south) -- (mh.south);
    \draw[decoration={brace, mirror}, decorate] (a1.south) -- (an.south);

    \draw[->] (mlstmq) -- (hq);
    \draw[->] (question.east) -- ++(4mm,0) -- ++(0,10mm) --  (lstm1.west);

    \draw[->] (m1) -- (mlstm1);
    \draw[->] (m2) -- (mlstm2);
    \draw[->] (mh) -- (mlstmh);
    \draw[->] (question) -- (mlstmq);
    \draw[<->] (mlstm1) -- (mlstm2);
    \draw[<->] (mlstm2) -- (ellipsesm2);
    \draw[<->] (ellipsesm2) -- (mlstmh);
    \draw[<->] (mlstmh) -- (mlstmq);
    \draw[->] (mlstmq) -- (lstm1);
    
    \draw[->] (a1) -- (lstm1);
    \draw[->] (a2) -- (lstm2);
    \draw[->] (an) -- (lstmn);
    \draw[<->] (lstm1) -- (lstm2);
    \draw[<->] (lstm2) -- (ellipses2);
    \draw[<->] (ellipses2) -- (lstmn);

    \draw[->] (lstm1) -- ++(0mm,7mm);
    \draw[->] (lstm2) -- ++(0mm,7mm);
    \draw[->] (lstmn) -- ++(0mm,7mm);
    
    \draw[<-] (g1) -- ++(0mm,-7mm);
    \draw[<-] (g2) -- ++(0mm,-7mm);
    \draw[<-] (gn) -- ++(0mm,-7mm);
    
    \draw[->] (g0) -- (hg0);
    \draw[->] (g1) -- (hg1);
    \draw[->] (g2) -- (hg2);
    \draw[->] (gn) -- (hgn);
    
    \draw[->] (hq.north) -- ++(0,11mm) -|  ([xshift=-1mm] hg0.south);
    \draw[->] (hq.north) -- ++(0,11.5mm) -|  ([xshift=-1mm] hg1.south);
    \draw[->] (hq.north) -- ++(0,12mm) -|  ([xshift=-1mm] hg2.south);
    \draw[->] (hq.north) -- ++(0,12.5mm) -|  ([xshift=-1mm] hgn.south);
    
    \draw[->] (hg0) -- (softmax);
    \draw[->] (hg1) -- (softmax);
    \draw[->] (hg2) -- (softmax);
    \draw[->] (hgn) -- (softmax);
    
    \draw[->] (softmax) -- (p);

    \end{tikzpicture}

    \caption{Model architecture with all proposed encoding techniques. Here, $\mathbf{h}=
    \mathbf{h}(q)$ and $\mathbf{g}_i=\mathbf{g}(a_i, \mathbf{A}, q)$. The network is further conditioned on the conversational context.
    }
    \label{fig:model}
\end{figure}
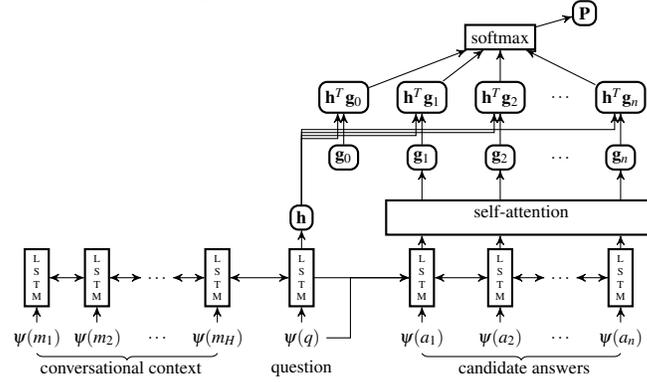

Our baseline model encodes $\psi(q)$ and $\psi(a_i)$ for $i=1,\:\ldots,\:n$, through feed-forward neural network layers. Other encoding techniques can also be introduced before the feed-forward layers, as in Figure \ref{fig:model}.
\vspace{-0.7cm}
\vspace{-0.6cm}

\begin{description}
	\item[\textbf{LSTM Layer}] \hfill \\
	The list of candidate answers $\mathbf{A}$ is an ordered sequence of sentences coming from the product description. Therefore, an LSTM could help add sequential information as context to the answer sentence representations \cite{lstm}. We specifically use a bi-directional LSTM and set the initial hidden state to be $\psi(q)$. This allows the question embedding to influence the answer sentence representations.
	\item[\textbf{Conversational Context}] \hfill \\
	Questions in our dataset occur in the context of chats. hence, the model could benefit from contextual information obtained from the messages before the question, $\mathbf{M} = \{m_1,\:m_2,\: \ldots,\: m_H\}$. We use an LSTM to model the embedded messages, taking the final LSTM hidden state as the question encoding \cite{lstm}.
	\item[\textbf{Attention Layer}] \hfill \\
	We use the feed-forward self-attention layer implemented in \href{https://github.com/tensorflow/tensor2tensor}{Tensor2Tensor} \cite{vaswani2017attention} to further enrich the representations of the candidate answers.
\end{description}

\subsection{Conversational Pre-training}

The full corpus of 36M conversations, containing 400M messages, was used to pre-train the model on the reply suggestion task \cite{henderson2017efficient}. The setup is similar to the baseline feed-forward model, except that the sentences to be ranked include the actual reply and messages randomly sampled from the corpus. The reply suggestion model achieved an 
\emph{Overall Accuracy} of 41.5\% on the answer sentence selection task.

\section{Evaluation Results}
\label{evalsec}
\begin{wraptable}{l}{0.63\textwidth}
\vspace{-0.7cm}
\begin{tabular}{ p{2.5cm} p{1.5cm} p{1.5cm} p{1.5cm}  }
 \toprule
  & \textbf{Overall Accuracy} & \textbf{Positive Accuracy} & \textbf{Trigger Accuracy} \\
 \midrule
 Baseline & 0.592 & 0.528 & 0.731\\
 + Pretraining & 0.678 & 0.604 & 0.783\\
 + LSTM Layer & 0.688 & 0.618 & 0.786\\
 + Attention  & 0.694 & 0.620 & 0.788\\

 + Conv. Context & 0.710 & 0.624 & 0.804\\
 \bottomrule
\end{tabular}
 \caption{Evaluation metrics for the different models.}
 \label{results}
 \vspace{-0.7cm}
\end{wraptable}
Table \ref{results} presents results for different model variations. Recall, from Section \ref{dataeval}, that overall accuracy is computed over all test samples while positive accuracy is computed over only samples with answers. Triggering accuracy evaluates how well the model predicts if there is an answer. All models contain 2 feed-forward or attention layers of size 500 for the baseline \& pre-trained models, 128 otherwise to reduce overfitting. N-gram embeddings and LSTM layers were of size 256. For the conversational context model, the message history is at most $H=10$ messages. Otherwise, the final 2 buyer messages are concatenated to form the question, $q$. The train-test split was ~90-10.

Pre-training on the full conversation corpus resulted in the greatest improvement, establishing the value of general conversational information. The improvement obtained from adding LSTMs over the candidate answers suggests that contextual information improves the answer sentence representations. Modelling the dialog state is also useful as evidenced by the improvement from using conversational context.
\vspace{-0.4cm}
% MH: small improvement from attention. improvement from conversation context shows the dialog state helps  x

% MH: can we make the question text larger in the figure? x
\section{QA Frontend}
\begin{wrapfigure}{l}{0.73\textwidth}
\vspace{-0.7cm}
\includegraphics[scale=0.24]{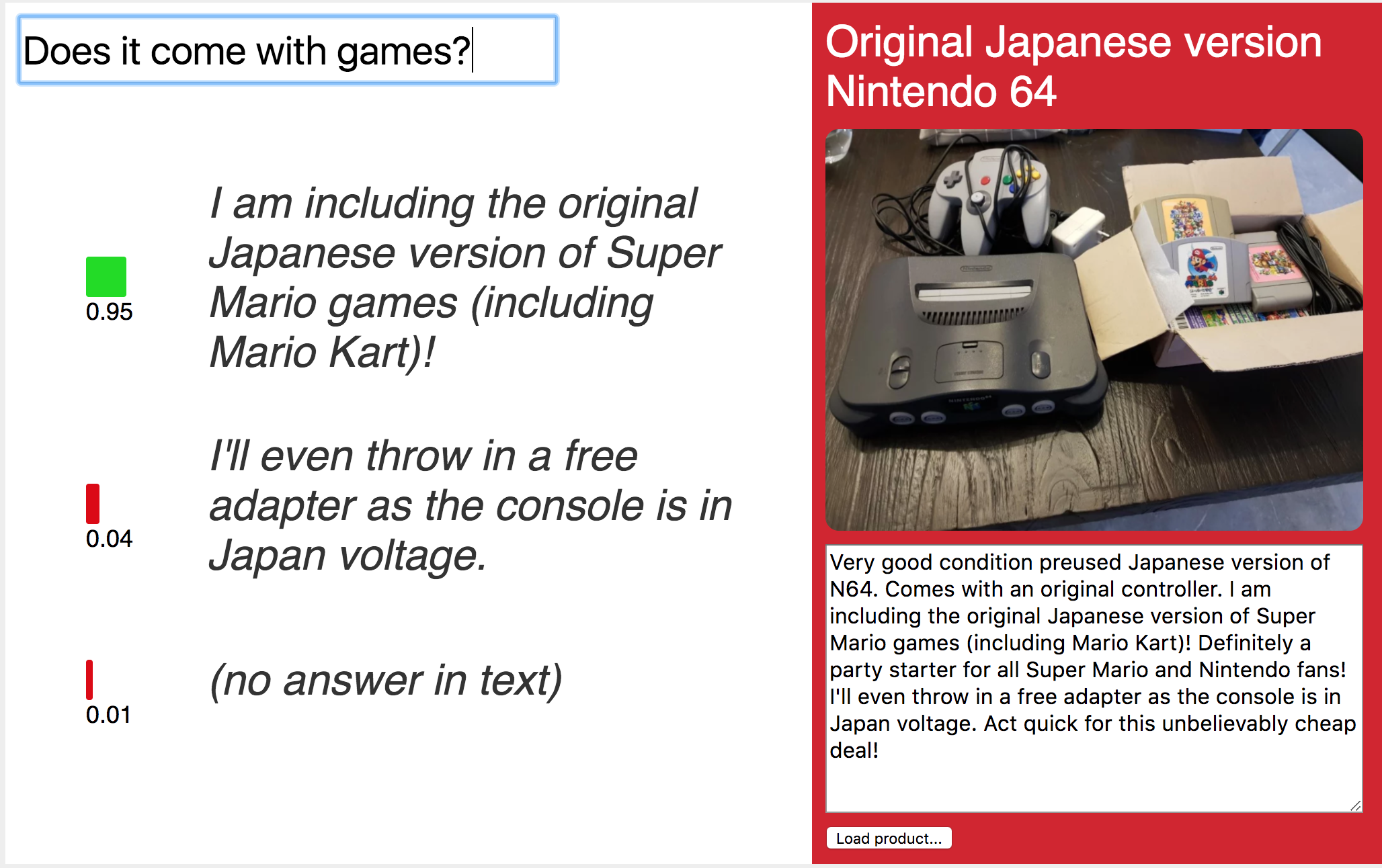}
\caption{Answer Selection Demo UI}
\label{qademo}
\vspace{-0.7cm}
\end{wrapfigure}
The intended use of this model is to present selected answer sentences as reply suggestions to sellers on \href{https://carousell.com}{Carousell}. For research purposes, we built a demo to facilitate exploring model predictions that will be presented at the International Workshop on Spoken Dialog System Technology, 2018.
The demo can import live product listings from \href{https://carousell.com}{Carousell}. When given a question, the demo presents sentences from the description that the model ranks as the best answers, including the likelihood of no answer being present (see figure \ref{qademo}).
The demo uses the best performing model trained on the final 2 buyer messages, i.e. Pre-training + LSTM + Attention. The conversational context model was not chosen for usability, as otherwise users would have to craft full conversations. 
\vspace{-0.4cm}
\section{Conclusion}
This paper has tackled the problem of answering buyer questions in online marketplaces using seller-crafted product descriptions. We first presented a neural-network ranking model for selecting sentences as answers. The introduction of a special no-answer token allowed the model to jointly decide whether an answer is present and to identify it if so. Multiple encoding techniques and a pre-training strategy were presented and evaluated. Finally, a demo was built to inspect model behaviour when answering questions about live products on the Carousell platform. The model performance was deemed good enough to be launched to power reply suggestions in Carousell. This means we can further fine-tune the model based on live user actions.

Future work could explore phrase-based question-answering methods to select more precise answers. It may also be interesting to study rephrasing the selected answer sentence to better fit the conversational context. Features like the listing title and product images could be introduced. We could also extend the system to answer questions without directly quotable answers in the description, e.g. yes/no questions. 

% MH: how about exploiting more features? e.g. title and image / user ID? x

% MH: extending to questions that do not directly quote the description, in training and inference- e.g. yes/no questions that can be answered by reading description x

\vspace{-0.4cm}

\bibliography{references}{}
\bibliographystyle{plain}
\end{document}